# Object Detection, Recognition, Deep Learning, and the Universal Law of Generalization

Faris B. Rustom[1], Haluk Öğmen[2], Arash Yazdanbakhsh[1]

[1] Department of Psychological and Brain Sciences, Computational Neuroscience and Vision Lab, Center for Systems Neuroscience, Boston University, Boston, MA, USA
[2] Department of Electrical & Computer Engineering, Laboratory of Perceptual and Cognitive Dynamics, University of Denver, Denver, CO, USA

## ABSTRACT

Object detection and recognition are fundamental functions underlying the success of species. Because the appearance of an object exhibits a large variability, the brain has to group these different stimuli under the same object identity, a process of generalization. Does the process of generalization follow some general principles or is it an ad-hoc "bag-of-tricks"? The Universal Law of Generalization provided evidence that generalization follows similar properties across a variety of species and tasks. Here we test the hypothesis that the internal representations underlying generalization reflect the natural properties of object detection and recognition in our environment rather than the specifics of the system solving these problems. By training a deep-neural-network with images of "clear" and "camouflaged" animals, we found that, with a proper choice of category prototypes, the generalization functions are monotone decreasing, similar to the generalization functions of biological systems. Our findings support the hypothesis of the study.

## STATEMENT OF RELEVANCE

The goal of the study is to discover general principles of object detection and recognition. A significant step was taken by the "universal law of generalization" as a law pertaining to all biological systems. The insight was to analyze generalization, not in the stimulus space, but in the "psychological space" (i.e., internal representations) of the species. However, since neurophysiological techniques do not allow a direct observation of the psychological space, the method was indirect. Recently, deep-neural-networks reached performance levels comparable to those of biological systems. Since we can examine the internal representations of deep-neural-networks directly, our approach was to train a deep-neural-network and examine their internal representations. Our results suggest that the process of object detection and recognition is shaped by invariant characteristics of the ecological environment and is independent of the specifics of the organism or the machine. This finding provides a bridge between Gibsonian ecological-approach and information-processing approaches.

## INTRODUCTION

In a real-world setting, object detection and recognition are extremely complex processes because stimuli impinging on our senses (or on the sensors of a device) are rarely identical. We can recognize a friend by looking at their face but the appearance of their face changes due, for example, to our own movements or the movements of the friend. These movements bring forth different perspective views of the face. Similarly, changes in lighting induce different shading patterns. Thus, even though object identity remains invariant ("my friend"), the phenomenal appearance can change drastically. Hence, the brain has to group these different appearances under the same object identity, a process of *generalization.*

## Universal Law of Generalization (ULoG)

Does the process of generalization follow some general principles and rules or is it an ad-hoc "bag of tricks"? If the universe, or more specifically the environment around us, has some invariances that can be described by laws, then it may be possible that some regularities of the generalization function can be identified not only within the members of the same species but also across different species since they are operating within the same environment obeying the same physical laws. This observation motivated Shepard to seek a "Universal Law of Generalization (ULoG)"[1]. He observed that previous attempts to find invariant rules of generalization failed because they sought to characterize the process in terms of the physical properties of the stimulus. His key insight was to approach the problem not in the physical space where the stimulus resides, but rather in the "psychological space", where the *internal representations* of the stimuli reside. Let $p_{ij}$ represent the probability of a learned response to stimulus $i$, to be made to stimulus $j$. Shepard defined

$$g_{ij} \stackrel{\text{def}}{=} \left(\frac{p_{ij}p_{ji}}{p_{ii}p_{jj}}\right)^{\frac{1}{2}} \tag{1}$$

as an empirical measure of generalization and plotted this empirical measure of generalization as a function of distance $d_{ij}$ between items $i$ and $j$ in the *psychological* space using data from animals and humans and from experiments using various tasks (e.g., judgments of size, shape, and hue; see Fig. 1 in [1]. The distance between stimuli in the psychological space was determined by a Multi-Dimensional Scaling (MDS) approach using pairwise stimulus proximity, or similarity, measures. Pairwise similarity measures can be considered as "distances" between stimuli and an optimization function finds a spatial representation where these pairwise similarity (distance) constraints are satisfied. The generalization measure as a function of distance in the psychological space showed a general *monotonic-decreasing* trend that captures the data very well for a variety of species and tasks (Fig. 1 in[1]). Given the generality of this finding in terms of tasks and species, it is reasonable to assume that similar solutions emerged in different species because they all share the same physical world and the laws governing this world act as common constraints in object detection and recognition. In other words, the solutions to object detection and recognition problems can be viewed as outcomes of a process of convergent evolution. Hence, these solutions are not only relevant for understanding how biological systems solve these problems but also for designing artificial systems that are expected to operate within the same environments as biological systems.

The ULoG being one of the few theories in psychology to approach the status of a general scientific law, it attracted significant interest[2–4], in particular to find an "explanation" for the law, i.e. abstract principles, such as algorithmic information-distance, from which the law can be derived. Our focus in this work was not to propose an explanation for the law, but instead, it was to examine whether the internal representations for object detection and recognition found in biological systems also emerge when learning is carried out by an artificial intelligence approach, more specifically by using deep-learning neural networks. It is well known that nonlinear feed-forward neural networks with three or more layers are "universal approximators", i.e., with the proper architecture and the dataset, they can learn *any* pattern detection and recognition problem with *any desired accuracy* [5–7]. Although some of these networks are inspired by biological systems[8], their architecture, neuron models, and operational and learning principles are too simple compared to human brains. However, if the internal representations observed by the ULoG hold across multiple species with significantly different neural and cognitive properties, and given the universal-approximation capability of artificial neural-networks, one can hypothesize that the internal representations that emerge reflect the natural properties of object detection and recognition in our environment rather than the specifics of the species solving these problems.

To test this hypothesis, we trained deep neural-networks and examined their internal representations to determine whether they possess properties similar to those of biological systems as analyzed by the ULoG. Moreover, to capture better the natural constraints of object detection and recognition, we used a “natural approach” to understand how evolutionary constraints shape internal representations: Interactions within and between species in nature exhibit strategies to promote or hinder detection and recognition. For example, predator-prey relations implicate mutual deception: the prey tries to deceive the predator to escape attack, whereas the predator deceives the prey to execute a surprise attack. In other words, camouflage is nature’s testbed for the robustness of object detection and recognition functions. Hence, we used camouflage in our analysis of internal representations to capture better the ecology of object detection and recognition and hence constrain further the possible solutions.

## Camouflage

Over millions of years, species developed a variety of camouflage forms. For more than two centuries, scientists have been observing and studying camouflage in nature. Notwithstanding the rich variety of species and their appearance, the general types of camouflage found in nature can be classified into few canonical types[9]: background matching, counter shading, self-shadow concealment, disruptive coloration, distractive markings, transparency, silvering, motion camouflage, and masquerade/mimicry. Of these categories, the large majority of samples in our dataset were background matching (75%) and masquerade/mimicry (15%). Background matching is a strategy to avoid detection by blending with the background, whereas masquerade/mimicry refers to a set of strategies aimed at disrupting recognition. Here the animal can be easily detected; however, features of the animal are modified so as to appear like an object of no interest (masquerade) or like a prey harmful to the predator (mimicry). As mentioned in the previous section, the ULoG proposes that the probability of two stimuli to be judged as being members of the same class or category (generalization) is a monotonically decaying function of their distance in the psychological space. Psychological space refers to how stimuli are *represented internally* by the animal. We use the term “feature spaces” in an inter-changeable way to highlight dimensions of these spaces as features that are extracted and analyzed. Within the context of camouflage, the animal has to generalize specific stimuli into broad categories of prey or predator animals as opposed to other items in the scene. Hence, camouflage consists of minimizing the probability of being classified as a prey or predator. In this context, similarity could be with respect to a predator, a prey, or a visual background such as a leaf pattern, a tree bark, etc.

According to ULoG, the probabilities of successful camouflage and camouflage-breaking strategies can be viewed as modifications of appearance in feature spaces. From a functional point of view, these modifications are intended to shift away from “detectable” feature sub-spaces to non-detectable regions to avoid detection, i.e., to systematically minimize the probability of detection/recognition. Alternatively, these modifications can also invade sub-spaces representing other objects so that the animal that undergoes camouflage is masquerading as an uninteresting object (e.g., insect appearing as a stick) or mimicking the appearance of a harmful prey. In this case, the strategy corresponds to shifting the appearance in the feature space to maximize the probability of being recognized as *another* object.

Shephard’s approach was to start with animal or human behavioral data on gradient of generalization (for example train an animal by reinforcing key-pecking response for a specific hue and test the frequency of pecking responses to other hues), and then to apply multi-dimensional scaling (MDS) to determine the underlying psychological-space distances and the associated gradient of generalization function. This is because, even with most advanced electrophysiological or imaging techniques, it is not possible to “look at” the internal representations in the brain and to infer psychological distances. In our work, instead of a biological system, we used a convolutional neural network (CNN) trained on an animal recognition task using both clear and camouflaged animal pictures. Unlike biological systems, we can directly access all details of internal representations that emerge in the artificial system and hence test directly whether or not the generalization-gradients that emerge follow the ULoG.

## METHODS

AlexNet[10] was the convolutional neural network (CNN) used in our study, implemented through MATLAB 2019b, with both training and testing performed on images of animals that were either in camouflage or clearly visible. The first source of images that was used was a public dataset being used for a similar camouflage detection project[15–17]. This dataset consisted of 2,500 total images, composed of predominantly camouflage images and some clear images. We supplemented the data with hundreds of additional images from Google Images to balance the number available in each category. The final dataset used for training and testing consisted of 3019 total images (1825 camouflage, 1194 clear), divided amongst 15 animal classes - Bear, Bird, Bulky Insect, Canine, Feline, Flat Fish, Flat Insect, Frog, Horse Type, Octopus, Owl, Reptile, Small Fish, Small Mammal, Stick Insect. The camouflage dataset featured examples of different types of camouflage but mainly consisted of animals disguised in background matching and mimicry camouflage techniques, comprising about 75% and 15% of the dataset respectively. The remaining camouflage types weren't as prominent in the dataset, collectively making up 10% of the total images.

The images within each category were randomly split into training (70%) and testing (30%) sets, and resized to 227 x 227 pixels to fit the AlexNet input requirement. Four total networks were trained on the images. The two direct networks are ClearNet, trained exclusively on clear images, and CamoNet, trained exclusively on camouflage images. The other two networks involved a transfer-learning step resulting in them being trained on both image types in different orders. Thus, ExpCamoNet was trained on clear images and transfer-learned onto camouflage images, while ExpClearNet was trained on camo images first and then transfer-learned onto clear images. All networks were trained on the same parameters, using a SGDM optimizer with an Initial Learning Rate of 0.001, a Learning Rate Drop Factor of 0.05, a Validation Frequency of 10, for 100 Epochs with training set shuffling to occur every Epoch. The 23$^{rd}$ layer, containing the third and final fully connected (FC) layer of AlexNet, was modified from the default 1,000 nodes to 15 nodes reflecting the number of animal categories. Training was performed on a single NVIDIA GeForce GTX 1050 Ti GPU.

In order to map the feature spaces of the networks, we extracted the 15 node activations of the category layer, each node belonging to a category (i.e. bear, frog, etc.) and are within the fully-connected (FC) layer from each network. We reduced the 15-dimensional activation to a 3-dimensional array via principal component analysis (PCA). The 3-dimensional activations of this array were plotted across the first three Principal Components (PC 1-3) to represent the networks' feature space in three dimensions. The feature spaces of each network were visually compared to identify patterns in cluster spreading and overlapping among the different networks. In these feature spaces, each point within the clusters represents the activation resulting from a single testing image on the trained FC layer. The position of each image's activation within the feature space is a visualization of the network's classification ability for each animal category cluster.

In addition to PCA, which is a linear method, we also used Multi-Dimensional Scaling (MDS). The Euclidean or the Manhattan distance between the activations generated by each image pair was calculated and passed to the MDS algorithm to obtain and plot the networks' feature space for the first three dimensions. Having *n* images, the total number of image-pairs is $n^2/2$.

The convex hulls were obtained using MathWorks convhull command which extracts the smallest convex region that contains the data points of a given species in the dimensionally-reduced space.

The animal clusters were further analyzed by plotting the images' corresponding category node activations as a function of the Euclidean and Manhattan distances between the image's point within the cluster and (1) the centroid or (2) the *true center*. The activations and centroid distance plots revealed a bifurcation of positive and negative slopes within individual animal groups indicating a disconnect in the relationship between the cluster centroid and the strength of activations of images immediately around it. In order to investigate whether a monotonically decreasing activation vs. distance plots can be obtained, we employed a search algorithm that sampled the feature space with a grid in the cluster to iteratively generate activation vs. distance plots for each grid-intersect as a center. This method allowed us to determine the "*true center*" of each cluster, which is the point from which calculating the distance between it and every other image produced a monotonically decreasing plot for activation as a function of distance.

## RESULTS

### Learning and generalization performance

The first step in our approach was to train a convolutional neural network for animal categorization task. We used AlexNet[10] as the "seed network". Our dataset was composed of a collection of images of camouflaged and "clear" animals. Some examples from the dataset are shown in the Supplementary Materials. We investigated four training approaches: train AlexNet on (i) clear images only, (ii) camouflaged images only, (iii) camouflaged images first followed by clear images, and (iv) clear images first followed by camouflaged images. We refer to the trained DNNs resulting from these four different training strategies as "ClearNet", "CamoNet", "ExpClearNet", and "ExpCamoNet", respectively. Tables 1a and 1b show the accuracies obtained by each network with the training (learning performance) and validation (generalization performance) datasets for clear and camouflaged images.

------ Table 1a and 1b-----

As can be seen from Table 1a, learning performance is consistently high across all training regimes with the exception of CamoNet's training accuracy on camouflage images. This discrepancy, however, is remedied following transfer learning. The generalization performance, assessed by accuracies obtained for test images (Table 1b), show that ClearNet has higher accuracy for clear images but lower accuracy for camouflaged images when compared to CamoNet. The two transfer-learned networks have virtually identical overall performance: They tend to produce a classification performance for clear images close to the performance of the ClearNet while simultaneously improving the performance for camouflaged images. In fact, these two networks performed better on camouflaged images than the CamoNet.

### Internal representations and the geometry of object recognition in feature spaces

Unlike biological systems, DNNs allow us to examine directly the internal representations that emerge from learning in order to investigate how they learned and how they generalize. The activities of neurons in the final fully-connected layer represent the ultimate geometric mapping of inputs for categorization. However, this layer has 15 neurons producing a 15-dimensional space. In order to visualize and analyze these representations, we used two dimensionality-reduction methods: One was the nonlinear method, the Multi-Dimensional Scaling (MDS), which was also used by Shepard, and the second the simpler widely-used linear method, Principal Component Analysis (PCA) (Fig. 1). Each image in the dataset is plotted as a point in the space with reduced-dimension. A first property that can be estimated from these geometric representations is the nature of the underlying metric. For example, Euclidean ($L_2$) and Manhattan ($L_1$) norms yield elliptic and rhombic iso-contours. Shepard suggested that, for simple stimuli such as colors, the metric should be Euclidean whereas for more complex stimuli, such as shapes, the metric should be Manhattan[1]. Our study is based on shape analysis and hence according to Shepard's approach, the Manhattan distance and the rhombic iso-contours would be expected. However, Shepard's analyses apply to the highly simplified learning/testing regimen that consists of a single learning trial followed immediately with a test trial. Nonetheless, Shepard also pointed out that for more general learning/testing regimens, wherein training includes several similar stimuli and testing is done with time delay, internal noise can transform rhombic iso-contours into elliptic ones[1]. It is difficult to estimate accurately the iso-contours in our case because of our dataset, hence the points in the feature space, do not provide a dense sampling of the space. However, we conducted an approximate analysis of the geometry of different regions in the feature space by estimating the convex hull bordering each region (Fig. 1). As we discuss below, in general we found elliptic contours, as one would expect from the type of learning we implemented in the DNN (animal categories based on shape). Below, we discuss the relationship between regions corresponding to different categories and their implications for learning and generalization. Highly overlapping regions indicate poor generalization. Supplementary Materials provide a comparison of different training and testing regimes. In summary, both ExpClearNet and ExpCamoNet training paradigms revealed an important property pertaining to the order in which transfer learning is performed. In each case, the networks achieved the highest accuracy on a particular task when most recently trained on that image type. Therefore, ExpClearNet was more accurate on clear images than ExpCamoNet (84.17% vs 79.61%, Table 1b), as clear images were the last set it was trained on. Likewise, ExpCamoNet was more accurate on camouflage images than ExpClearNet (64.88% vs 62.07%, Table 1b), since camouflage images were the final training set. This is likely due to the stability-plasticity dilemma[11] or catastrophic interference/forgetting[12] property of these types of feed-forward networks: Whereas the initial training "molds" the network, when the network is trained with a different dataset, the error minimization algorithm operates only on this new dataset thereby gradually eroding old memories.

----- FIG 1-----

For a more focused approach, we considered four animal classes split into two supergroups, where the animals within each supergroup are visually similar to one another, yet very different from the species belonging to the other supergroups. Specifically, the first supergroup consisted of Bear and Canine images while the second consisted of Frogs and Reptiles. The positions of these Bear/Canine and Frog/Reptile supergroup regions in a feature space across different principal components were impacted by transfer learning of the networks (see Supplementary Materials for other learning regimes), in a manner consistent with the effects of transfer learning for all animal activation regions. Transfer learning appeared to be most effective at resolving the confusion associated with a network's respective foreign[1] image dataset. For both ExpClearNet's/ ExpCamoNet's camouflage/clear image activation feature-spaces, when compared to the original ClearNet/CamoNet feature-spaces, respectively, transfer learning significantly reduced between-group mixing of the Bear/Canine and Frog/Reptile animal regions across the higher principal components 2 and 3 (compare Fig. 2 a-c and Fig S7 d-f with S5 a-c and S4 d-f, respectively), which were previously completely overlapping. In other words, the transfer learning step introduced to ClearNet and CamoNet (resulting in ExpClearNet and ExpCamoNet, respectively) was critical in resolving the confusion surrounding the complex, higher order features within the latter principal components that allowed the network to distinguish the two animal region supergroups. These transformations within the feature space demonstrate the improvement in network performance and accuracy on both clear and camouflage animal recognition tasks, represented by increased distinction between animal regions, as a direct result of transfer learning.

----- FIG 2-----

## Distances in feature spaces and the generalization functions

As mentioned before, Shepard's theory was based on an experimental setting wherein generalization was assessed with a novel stimulus delivered right after a single learning trial[1]. He suggested that the theory could be expanded to more realistic and general cases wherein training includes several similar stimuli and testing is done with time delay. He suggested that noise needs to be considered in this more general learning/testing regimen transforming the rhombic iso-contours into elliptical ones and changing the exponential decay function of generalization into a Gaussian function. Below we analyze the gradients of generalization that emerge from the CNN.

[1] We use the term "foreign" for datasets that are, not only absent from the training set, but also differ in terms of their nature, i.e., clear vs camouflaged images.

In Shepard's approach, there is a single learning trial and a single test stimulus. The learning and test stimuli are placed at the center of their region in the feature space and the test stimulus' probability of generalization is calculated according to the overlap between these two "consequential regions"[1]. Here, we have multiple similar stimuli for each category. With multiple stimuli, generalization is not based on individual comparisons between two stimuli but rather is likely to be based on a "prototype" that represents the category. In fact, some alternative learning theories, such as the Adaptive Resonance Theory[11] carry out the classification by comparing the input with category prototype that emerges from learning through feedback interactions (as opposed to purely feed-forward approach of the CNN). So, we investigated if we can identify such a prototype for each category that will lead to well-behaved generalization functions.

As a first step, we considered the centroid of the region as the prototype of the category and computed generalization functions. Fig 3 shows the results. The generalization functions show a bifurcation: The data points do not cluster along a single decreasing function but instead appear to cluster along two functions, one increasing and another one decreasing with distance. Hence, with respect to the centroid, the generalization becomes a multivalued function yielding two different predictions.

----- FIG 3-----

Next, we investigated whether a prototype point can be found to produce uni-valued generalization function (see Methods for the approach). In addition, we tested both linear (PCA) and nonlinear (MDS with Euclidean and Manhattan metrics) dimensionality reduction. Figure 4 shows the resulting feature spaces and Figure 5 plots the generalization functions that emerge from these *prototype* points, that we call the "true-center'. Although data still appear noisy, when we compare these generalization functions with that obtained with respect to the centroid (Fig. 4), first we see a single monotonic decay tendency, as opposed to the bifurcation in Fig. 4. Second, data cluster better along a decreasing linear trend. This is particularly true for the MDS Euclidean case (Fig. 5, middle row; see R-squared values for linear regression in the caption).

----- FIGs 4 and 5-----

To analyze to which extent these results reflect ecological settings and camouflage strategies, we plotted generalization functions of the ClearNet for clear and camouflaged animals (Fig. 6). ClearNet is trained exclusively with clear images and its performance for camouflaged images reflects its generalization ability for camouflaged inputs.

----- FIG 6-----

As it can be seen from Fig. 6, camouflaged inputs are not distributed randomly but also fall on the same generalization line as the clear images. With the exception of the tail of the canine generalization curve that seems to exhibit a floor effect, these functions are remarkably linear (see R-squared values for linear regression in the caption of Figs. 5-6). This suggests that learning is robust with respect to camouflage in that similar processes are used for both clear and camouflaged images of the same animal category. For ClearNet, training samples did not include camouflaged images and the "true center", i.e., the prototype, is computed based on these clear images. When camouflaged images' distance is calculated based on the prototype determined on clear images, instead of observing a random or a bifurcating tendency as in Fig. 3, we do find an orderly relation that follows the generalization function of clear images. Moreover, we also notice that camouflaged images tend to cluster at the bottom of this generalization curve, indicating that camouflage is successful in reducing the probability of detection and recognition.

These results show both similarities and differences between biological generalization curves and those that emerge from CNNs. In both cases, one can identify monotone decreasing generalization curves. Whereas CNN generalization curves tend to be linear, biological generalization curves are nonlinear (exponential or Gaussian).

## DISCUSSION

The fundamental hypothesis of our study was that if the internal representations observed by the ULoG hold across multiple species with significantly different neural and cognitive properties, and given the universal-approximation capability of artificial neural-networks, then the internal representations that emerge reflect the natural properties of object detection and recognition in our environment rather than the specifics of the species solving these problems. A similar hypothesis in visual perception stems from the frequency spectrum of natural images. It is known that natural images tend to exhibit a frequency spectrum with amplitude $A(f)$ that follows a power law: $A(f) \approx \frac{c}{f^{\gamma}}$ , where $f$ is spatial frequency, $c$ and $\gamma$ are positive constants[13]. It has been shown that visual neurons in a wide-range of species, from mammals[14] to insects[13] exhibit a frequency-tuning characteristic that is adapted to this power law, suggesting that despite strong differences between these species, a process of convergent evolution leads to similar functional characteristics in the sensory systems. This visual system example reflects a relatively "low level" neural computation whereas object recognition is considered a "high-level" neural computation. While the appearance of objects in the environment cannot be described by a simple power law which can be translated to frequency-tuning characteristics of individual neurons, we show here that a "universal generalization function" can be found by analyzing network (rather than individual neuron) representations at higher levels of processing hierarchy.

One major shortcoming of DNNs has been their opacity, i.e., it is difficult to know how they make their decisions. Our approach also provides a way to look "under the hood of DNNs". Our approach provides a principled method for identifying prototypes for different categories and explain the generalization behavior of the DNN according to distances in the reduced feature-space.

Finally, our findings can be viewed as a bridge between Gibsonian ecological approach and information processing approaches. We provide evidence that object detection and recognition is shaped by invariant characteristics of the ecological environment and is independent of the specifics of the organism or the machine, provided that they have the requisite power (cf. universal approximation theorem).

**Abbreviations:** CNN: Convolutional neural network; MDS: Multi-dimensional scaling; PCA: Principal component analysis; ULoG: Universal law of generalization.
**Data and code availability:** The dataset and the code used in this study are available at: https://www.kaggle.com/datasets/farisrustom/camoanimals?select=Code

## AUTHOR CONTRIBUTIONS

F.B.R, H.O. and A.Y. conceived the study. F.B.R and A.Y. carried out the training of the neural network. F.B.R, H.O. and A.Y. analyzed and interpreted the results. H.O. wrote the first draft of the manuscript. F.B.R and A.Y. edited the manuscript.

## DECLARATION OF INTERESTS

The authors declare no competing interests.

## SUPPLEMENTARY INFORMATION

### Some Examples from our Dataset

Fig. S1 shows some examples from the dataset we used. S1 a-b show clear vs. background matching, and c-d show clear and masquerade/mimicry, respectively.

------Fig. S1------

### Internal Representations and the Geometry of Object Recognition in Feature Spaces Resulting from Different Training Regimes

Noticeable transformations in dimension-reduced feature spaces were observed as a result of varying the training regime and the testing image dataset type. For ClearNet, the variance explained for the first six components are listed in Table 2, along with the calculated difference between subsequent component variances. Given the distinct drop-off in the difference of variance values between principal components three and four we can use the first three components to provide a good approximation for the geometry of internal representations. ClearNet demonstrated well-spread regions within its feature space on its native clear image dataset (Fig S2 a-d), in which the animal regions maintained their distinction and verged away from the center. On the other hand, the same ClearNet presented highly condensed and blended animal regions in its feature space on the foreign camouflage image dataset (Fig S2 e-h), where the animal regions were significantly mixed and converged towards the center. This pattern holds true across all of the first three PCs (Fig S2 b-d, f-h), demonstrating consistent confusion by the network on increasingly complex features abstracted from the images. The contrast between these feature spaces for the same network visually reflect ClearNet's disparity in accuracy on the two classification tasks (83.29% on clear vs 52.13% on camouflage, Table 1).

------Table 2------

------Fig. S2------

Similar observations were made regarding the CamoNet feature space, which demonstrated less central clustering of its regions on its native dataset of camouflage animal images (Fig S3 f-h) than ClearNet did on the same image type. However, when comparing the feature spaces of both networks on their respective native datasets, CamoNet's region separations are not as visible as ClearNet's and generally show increased blending among regions, indicative of increased network confusion (verified by an accuracy of 59.18%, Table 1b). This is an outcome of the inherently more challenging task camouflage animal images pose to the network, with the resulting lower network accuracy being reflected in the feature space clustering. CamoNet's foreign data feature space of clear images is also more blended and condensed than it was for its native camouflage data (Fig S3 b-d), following the same trend as ClearNet in which feature spaces of the networks' respective native dataset exhibit greater region distinction. Comparing both networks' foreign dataset feature spaces, we observe that CamoNet maintains clearer distinctions between regions than ClearNet manages to do (Table 1b). This again is likely an additional benefit CamoNet gains from being trained on the more challenging camouflage image dataset, such that the network struggles less when presented with the foreign, albeit simpler, clear-animal image-set.

------Fig. S3------

Introducing an intermediate transfer learning step to the training pattern of both network types produced notable transformations in the animal regions. Following transfer learning for ClearNet (i.e., initially trained on camo transfer trained on clear), the new ExpClearNet exhibited significantly less central clustering of the animal regions for both clear and camouflage animal activations (Fig S4). The accuracy of ExpClearNet on the clear animal task changes minimally, implying that the changes in layer weights accompanying transfer learning from camouflage animals has a relatively small effect on the network's accuracy on the clear animal task, despite changes in the corresponding feature space (Fig S4 a-d). The main effect of the transfer learning step, however, is best observed on the camouflage condition (Fig S4 e-h). ExpClearNet not only shows less central cluttering of the animal regions and increased region expansion (reflected by the increase to 62.07% accuracy from 52.13%, Table 1), but the feature space also appears to rotate within the three-dimensional space with animal regions being represented in different locations than prior to transfer learning, while still maintaining the relative positions of each animal regions to one another based on feature differences.

------Fig. S4------

While transfer learning mainly improved ExpClearNet's feature space distinction on the foreign camouflage dataset, ExpCamoNet's feature space benefitted in both its respective native camouflage and originally foreign clear animal image set (Figure 1, main text). The previously shown highly mixed CamoNet clear-image feature-space was significantly resolved following transfer learning, with animal regions migrating away from the center to appear better separated and more distinct (Fig 1 a-d), now rivaling the performance of both the original ClearNet and ExpClearNet with an accuracy of 79.61% (Table 1b) on this task. A similar effect is observed with ExpCamoNet's performance on camouflage images as well. Transfer learning generated the desired animal region expansions and migrations away from the center of the feature space (Fig 1 e-h), indicating that the newly trained network has gained a greater aptitude for camouflage animal classification through transfer learning (again reflected by a jump in accuracy from 59.18% to 64.88%, Table 1b).

Taking a more focused approach and only examining a few, carefully selected animal clusters at a time allows the feature space to be analyzed in greater depth, and the relationships between animal regions can be further examined. We considered four animal classes split into two supergroups, where the animals within each supergroup are visually similar to one another, yet very different from the species belonging to the other supergroup. Specifically, the first supergroup consisted of Bear and Canine images while the second consisted of Frogs and Reptiles. This visual comparison of the animal groups translates well into a feature space, as demonstrated by both ClearNet and CamoNet, where the networks identify differing species as such and separate their corresponding regions across the first two principal components, while the images of similar species pose a greater challenge to the networks and are thus blended together (Fig S5a, Fig S6a). The separation of the two supergroups is seemingly affected by the image type being shown to the network – that is, clear or camouflage images – since the increased network confusion caused by camouflage images results in reduced between-group separation, indicating a reduced distinctive ability by the network due to the camouflage effect (Fig S5d, Fig S6d).

------Fig. S5------

The degree of between- and within-group separation for the Bear/Canine and Frog/Reptile animal region supergroups was found to vary across different principal components. As mentioned previously, the ClearNet feature space of its native clear image dataset demonstrated notable between-group separation on early principal components (Fig S5a). Specifically, the Bear/Canine supergroup is completely separated from the Frog/Reptile supergroup across PC1 and maintains a relatively good separation between groups on PC2 and PC3 (Fig S5 b-c). This is again a reflection of ClearNet's performance on the clear image task. On the more challenging camouflage task, ClearNet's PC1 separation is significantly reduced (Fig S5d), and all animal regions become heavily mixed on PC2-3 (Fig S5 e-f). In other words, ClearNet can only base its animal classifications on features stored in PC1 following the introduction of camouflage images, which appear to affect ClearNet's ability to distinguish features stored in latter principal components. This is again observed in CamoNet's foreign clear image set feature space, where the regions of the two supergroups are well separated across PC1 but heavily mixed on PC2-3 (Fig S6 a-c). This suggests that in general, the more challenging features with which the network struggles are stored in latter principal components while lower-level features are stored within early principal components.

------Fig. S6------

The positions of these Bear/Canine and Frog/Reptile supergroup regions in a feature space across different principal components were also impacted by transfer learning of the networks, in a manner consistent with the earlier discussion of transfer learning effects for all animal activation regions for each network. Transfer learning appeared to be most effective at resolving the confusion associated with a network's respective foreign image dataset. In the case of ExpClearNet's camouflage image activation feature space compared to the original ClearNet, transfer learning significantly reduced between-group mixing of the Bear/Canine and Frog/Reptile animal regions across the higher principal components 2 and 3 (Fig S7 d-f), which were previously completely overlapping (Fig S5 e-f). In other words, the transfer learning step introduced to ClearNet (resulting in ExpClearNet) was critical in resolving the confusion surrounding the complex, higher order features within the latter principal components that allowed the network to distinguish the two animal region supergroups. The same effect was also observed with ExpCamoNet's clear image activation feature space, where once again the animal region supergroups (Bear/Canine as one, Frog/Reptile as the other) could only be distinguished by the network through transfer learning (Fig 2 a-c, main text). These transformations within the feature space demonstrate the improvement in network performance and accuracy on both clear and camouflage animal recognition tasks, represented by increased distinction between animal regions, as a direct result of transfer learning.

------Fig. S7------

Finally, Fig. S8 provides a comparison between generalization functions obtained by CamoNet and ExpCamoNet using MDS. Both networks show similar monotone decreasing generalization functions. Inspecting the far-right panels that plot together the generalization functions for different categories, we can observe that ExpCamoNet ‘s generalization functions appear to show vertical separation for different categories whereas those of the CamoNet tend to be clustered together. This indicates that different categories have different distance-generalization properties for ExpCamoNet but similar properties for CamoNet. In other words, for ExpCamoNet, the “tightness” of the group varies by category but it is relatively the same for CamoNet.

# TABLES

Table 1a. Network training accuracy on image types

| Network | Trained On | Mean Accuracy (%) |
| --- | --- | --- |
| ClearNet | Clear | 98.23 |
| CamoNet | Camo | 87.03 |
| ExpClearNet | Camo → Clear | 99.27 |
| ExpCamoNet | Clear → Camo | 98.80 |

Table 1b. Network testing accuracy on image types

| Network | Trained On | Tested On | Mean Accuracy (%) |
| --- | --- | --- | --- |
| ClearNet | Clear | Clear | 83.29 |
| ClearNet | Clear | Camo | 52.13 |
| CamoNet | Camo | Clear | 71.49 |
| CamoNet | Camo | Camo | 59.18 |
| ExpClearNet | Camo → Clear | Clear | 84.17 |
| ExpClearNet | Camo → Clear | Camo | 62.07 |
| ExpCamoNet | Clear → Camo | Clear | 79.61 |
| ExpCamoNet | Clear → Camo | Camo | 64.88 |

Table 2. PCA variance and difference values for trained networks

| ClearNet | | CamoNet | | ExpCamoNet | |
| --- | --- | --- | --- | --- | --- |
| Variance | Difference | Variance | Difference | Variance | Difference |
| 32.6337 | | 64.0825 | | 51.0757 | |
| 23.3585 | -9.2752 | 50.7853 | -13.2972 | 38.4322 | -12.6435 |
| 9.5994 | -13.7592 | 19.2611 | -31.5242 | 16.1704 | -22.2619 |
| 6.8957 | -2.7036 | 13.3533 | -5.9078 | 13.1592 | -3.0112 |
| 5.3606 | -1.5351 | 12.1472 | -1.2061 | 12.3314 | -0.8277 |
| 4.2050 | -1.1556 | 10.8898 | -1.2574 | 10.1001 | -2.2313 |

# FIGURES

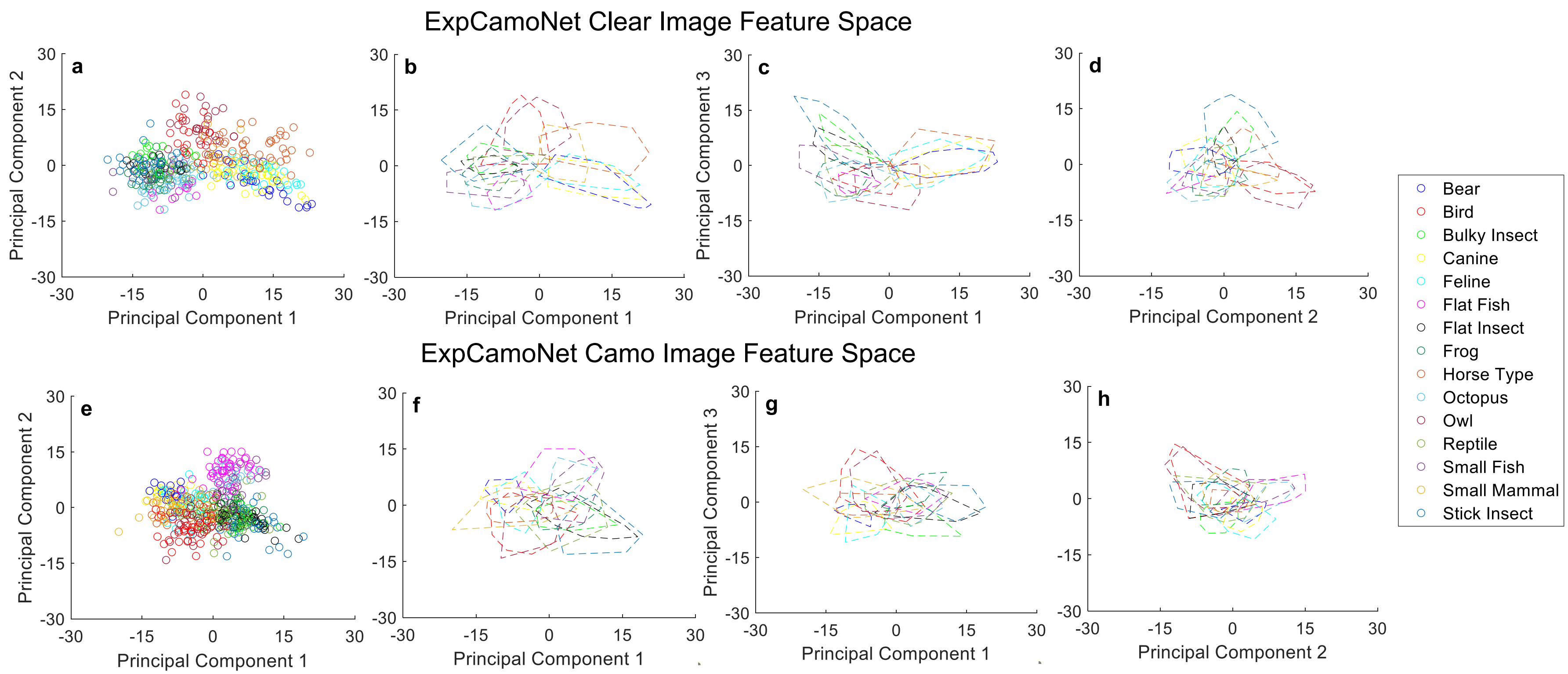

**Figure 1 –** PCA-based feature-space comparison between ExpCamoNet activations on clear and camouflage animal testing images across the first three principal components**. (a)** Cluster representation of clear images on XY axis. **(b-d)** Region representation of clear images on XY, XZ, and YZ axes. **(e)** Cluster representation of camouflage images on XY axis. **(f-h)** Region representation of camouflage images on XY, XZ, and YZ axes. Contours in b-c & f-h are convex hulls surrounding each species data points in a & e respectively.

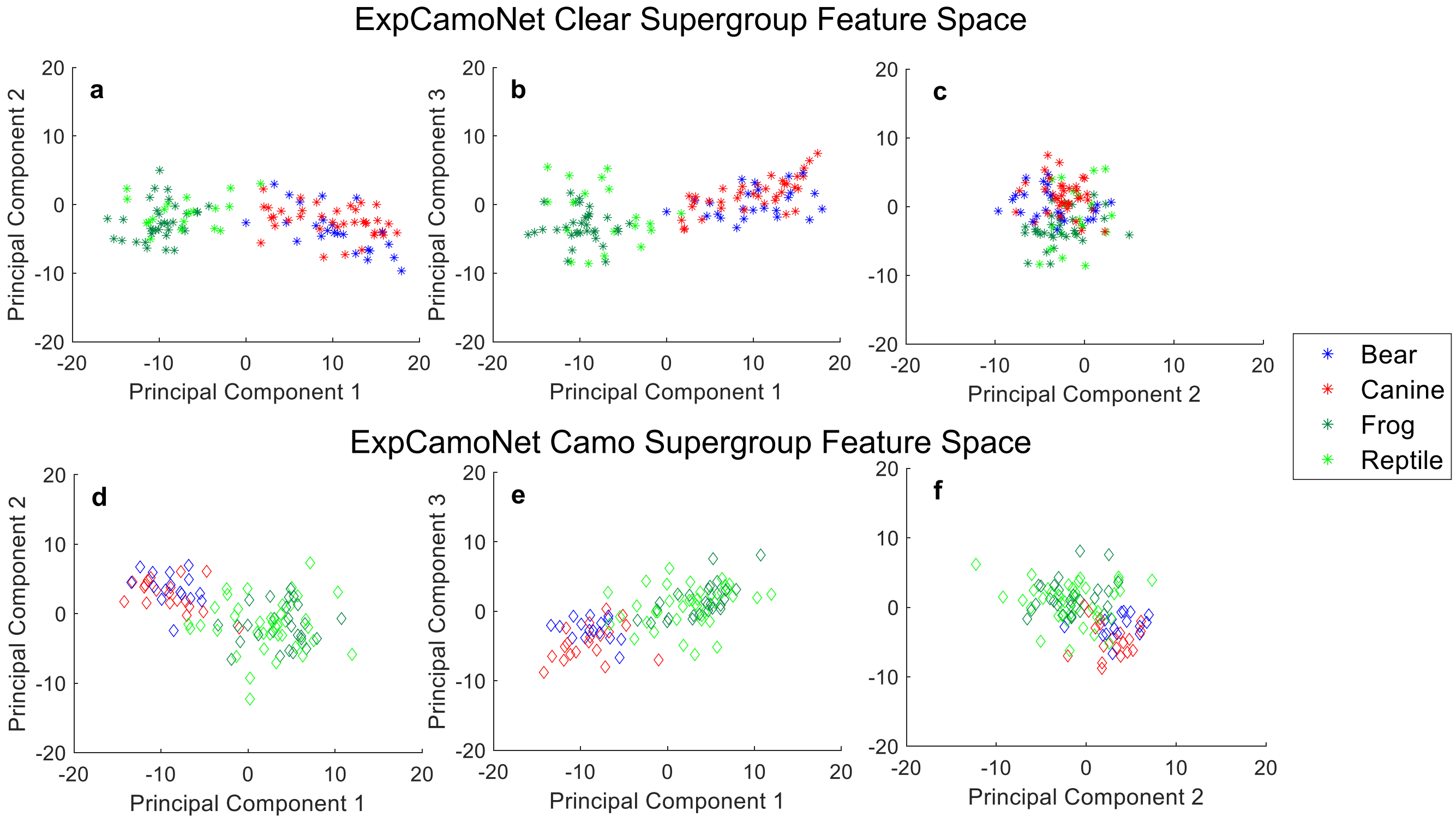

**Figure 2** – PCA-based feature space comparison between ExpCamoNet activations on select clear and camouflage animal "supergroup" images across the first three principal components**. (a-c)** Cluster representation of clear images on XY, XZ, and YZ axes. **(d-f)** Cluster representation of camouflage images on XY, XZ, and YZ axes.

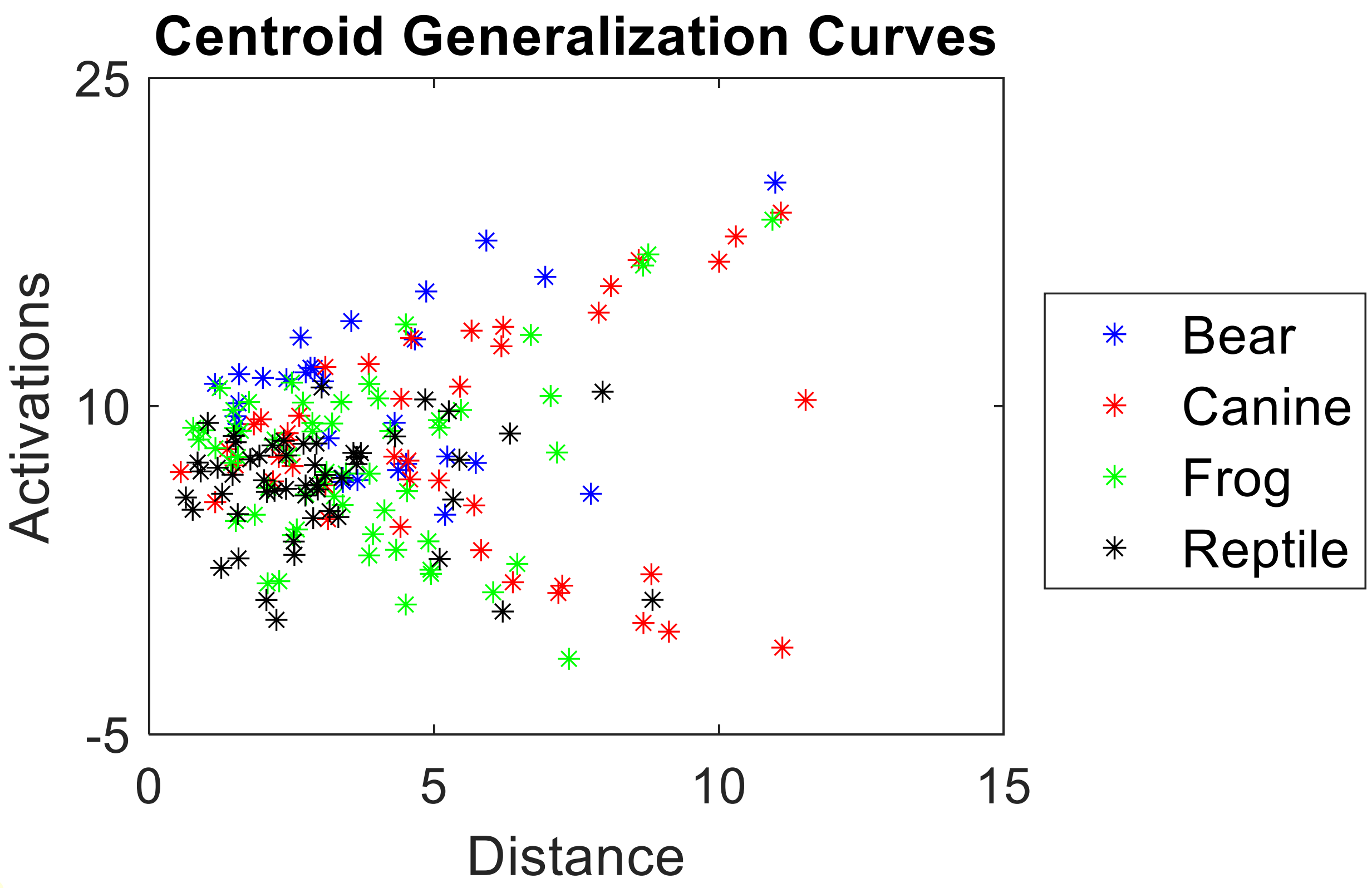

**Figure 3** – Generalization functions computed with respect to the centroid of the region in the feature space for the ExpCamoNet. The data points do not cluster along a single decreasing function but instead appear to cluster along two functions, one increasing and another one decreasing with distance.

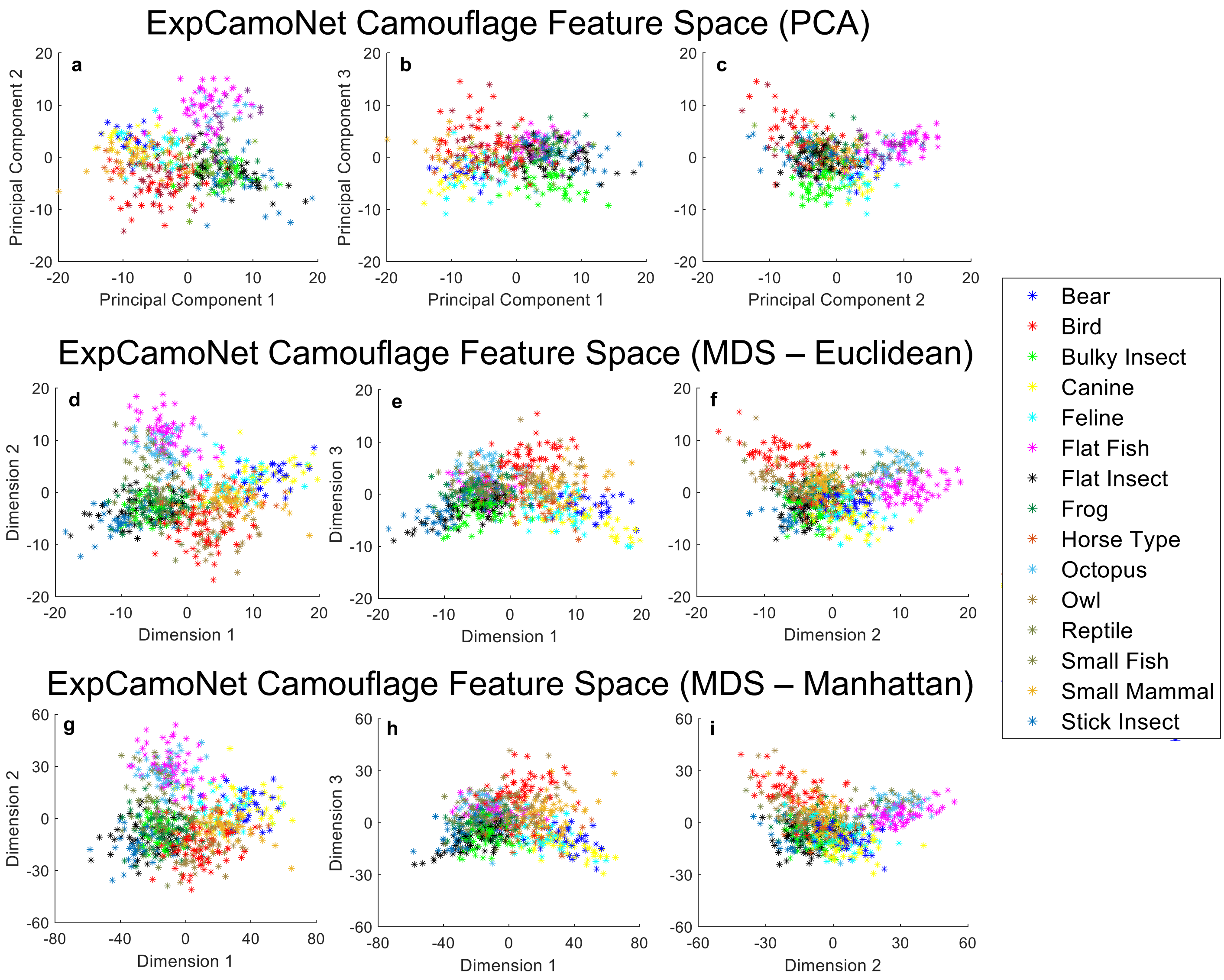

**Figure 4** – Feature space comparison between two dimensionality reduction methods (PCA and MDS) for ExpCamoNet activations on camouflage animal test images across three dimensions. The cluster representation of camouflage animals are shown when generated using PCA **(a-c)**, MDS with a Euclidean distance calculation **(d-f)**, and MDS with a Manhattan distance calculation **(g-i)**.

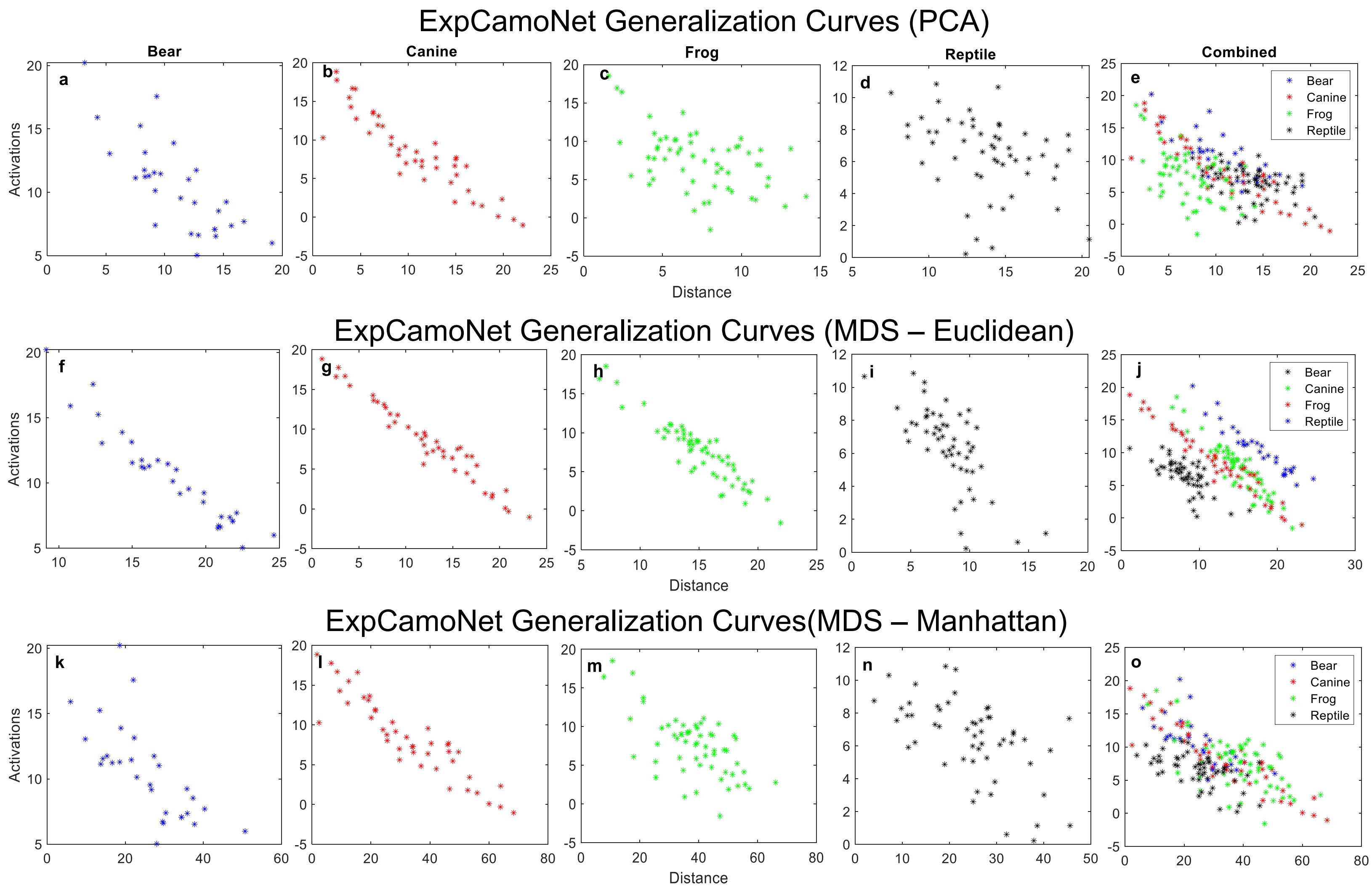

**Figure 5** – Comparison of ExpCamoNet network activation as a function of distance between Principal Component Analysis (PCA) **(top row)** and Multidimensional Scaling (MDS) with Euclidean **(middle row)** and Manhattan **(bottom row)** distance calculation. Four animal categories for each network are shown individually – Bear, Canine, Frog, and Reptile, as well as together. R-squared values for linear regression: a) 0.6037 b) 0.8317 c) 0.2511 d) 0.1242 f) 0.9229 g) 0.9515 h) 0.8582 i) 0.5005

k) 0.5100 l) 0.8390 m) 0.4191 n) 0.3291.

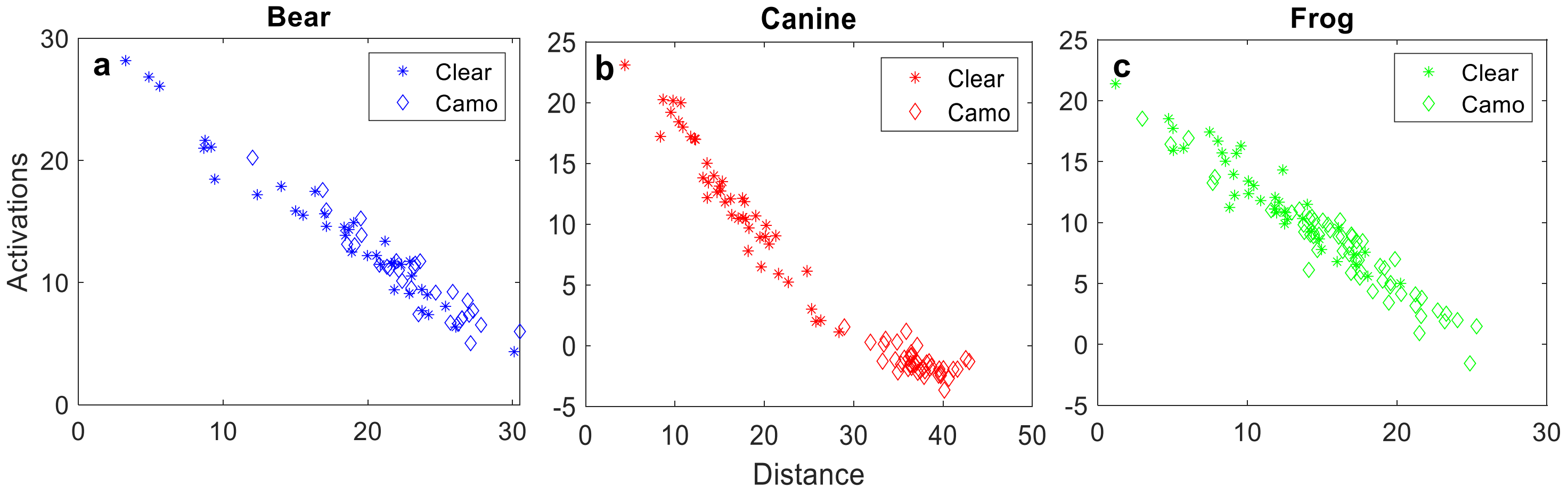

**Figure 6 –** ClearNet generalization curves for clear and camo animals of each group, relative to the clear cluster true center. The true center (the prototype) of the clear animal group in each case was found and used to measure the distance between the clear and camo images. Uses activation values from the Fully Connected Layer (23rd layer). R-squared values for linear regression: a) Clear – 0.9579, Camo – 0.8940; b) Clear – 0.9456, Camo – 0.4270; c) Clear – 0.8929, Camo – 0.9096.

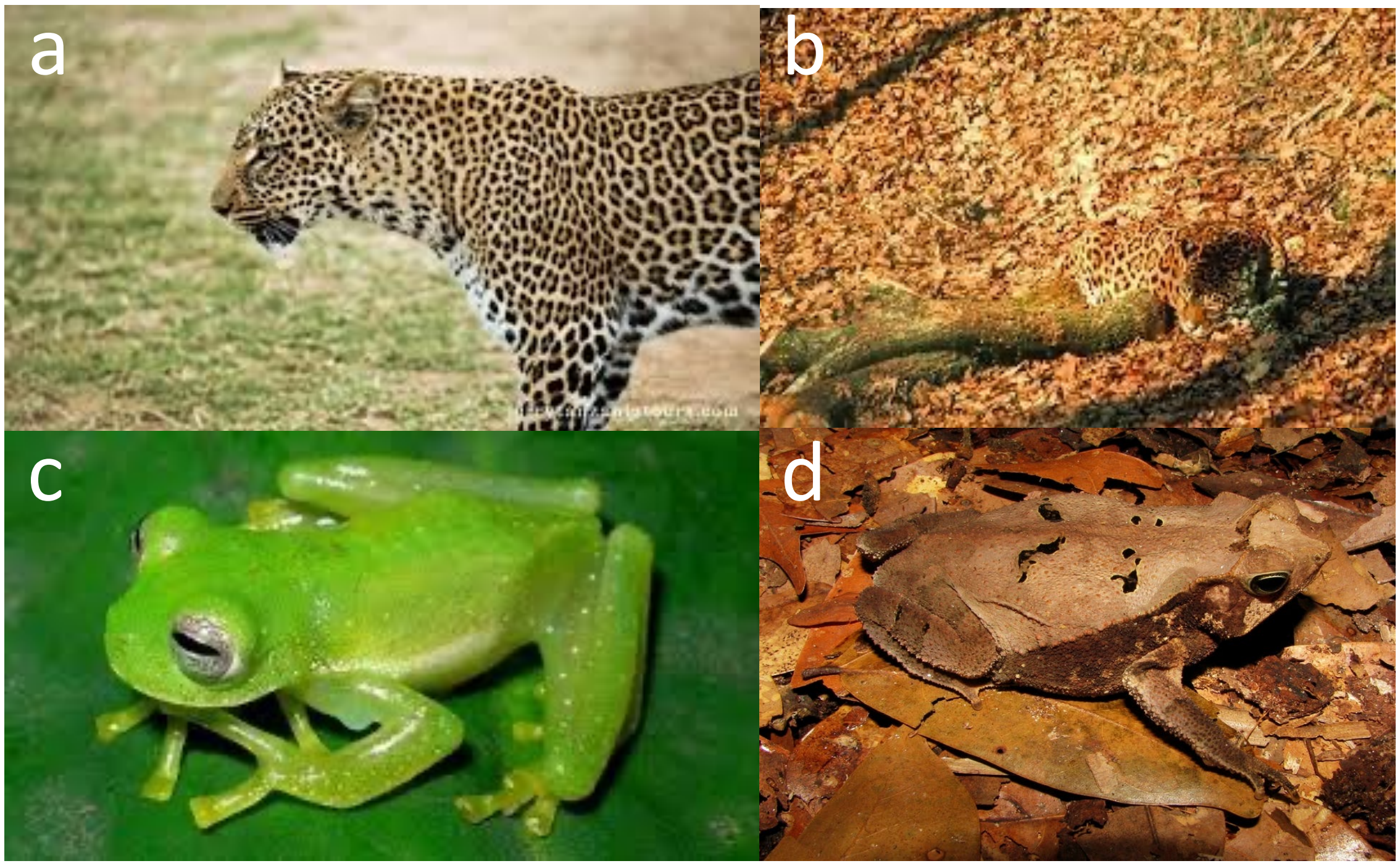

**Figure S1 –** Some examples of images from the dataset used to train and test the network. a) clear animal and b) in camouflage based on background matching, c) clear and b) masquerade/mimicry-based camouflage (adopted from references 15-17).

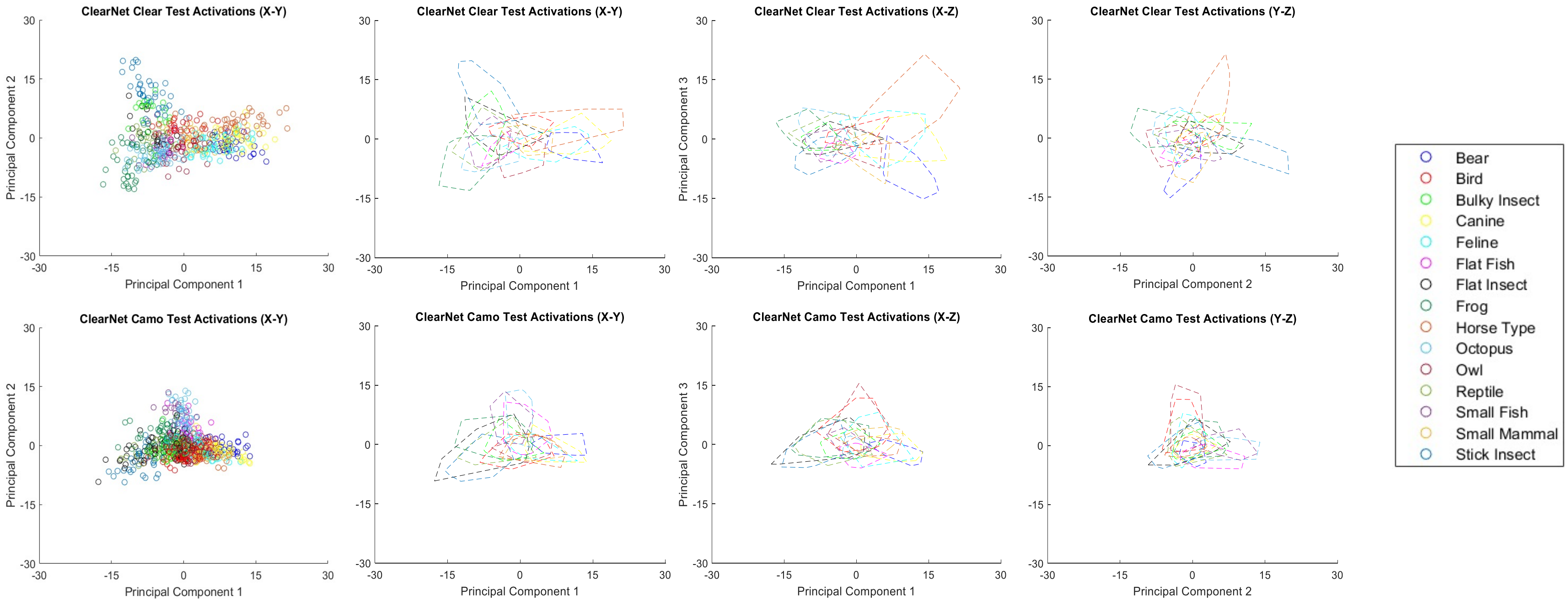

**Figure S2 –** Feature space comparison between ClearNet activations on clear and camouflage animal testing images across the first three principal components, which are reflected on the x-,y-, and z-axis, respectively**.** The Convex Hull around each cluster is used to illustrate each species' *region*. **(A)** Cluster representation of clear images on XY axis. **(B-D)** Region representation of clear images on XY, XZ, and YZ axes. **(E)** Cluster representation of camouflage images on XY axis. **(F-H)** Region representation of camouflage images on XY, XZ, and YZ axes.

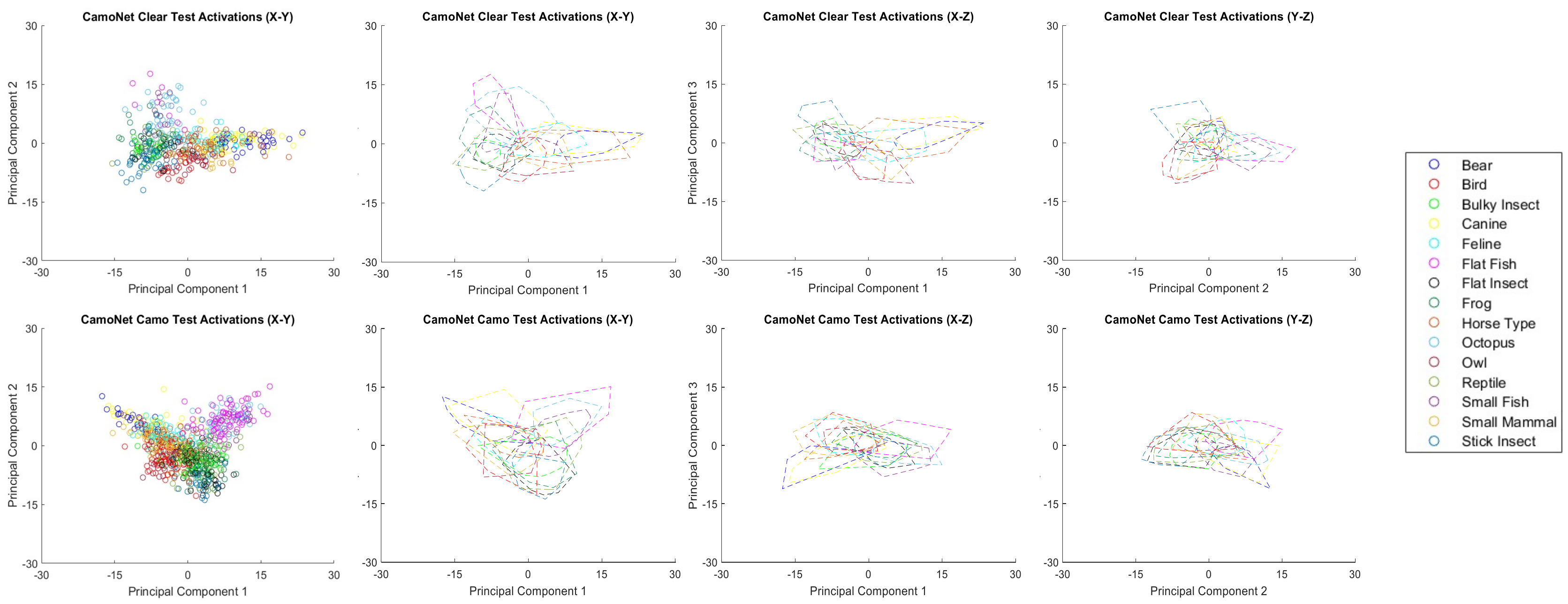

**Figure S3** – Feature space comparison between CamoNet activations on clear and camouflage animal testing images across the first three principal components**.** Conventions are as in Figure 1. **(A)** Cluster representation of clear images on XY axis. **(B-D)** Region representation of clear images on XY, XZ, and YZ axes. **(E)** Cluster representation of camouflage images on XY axis. **(F-H)** Region representation of camouflage images on XY, XZ, and YZ axes.

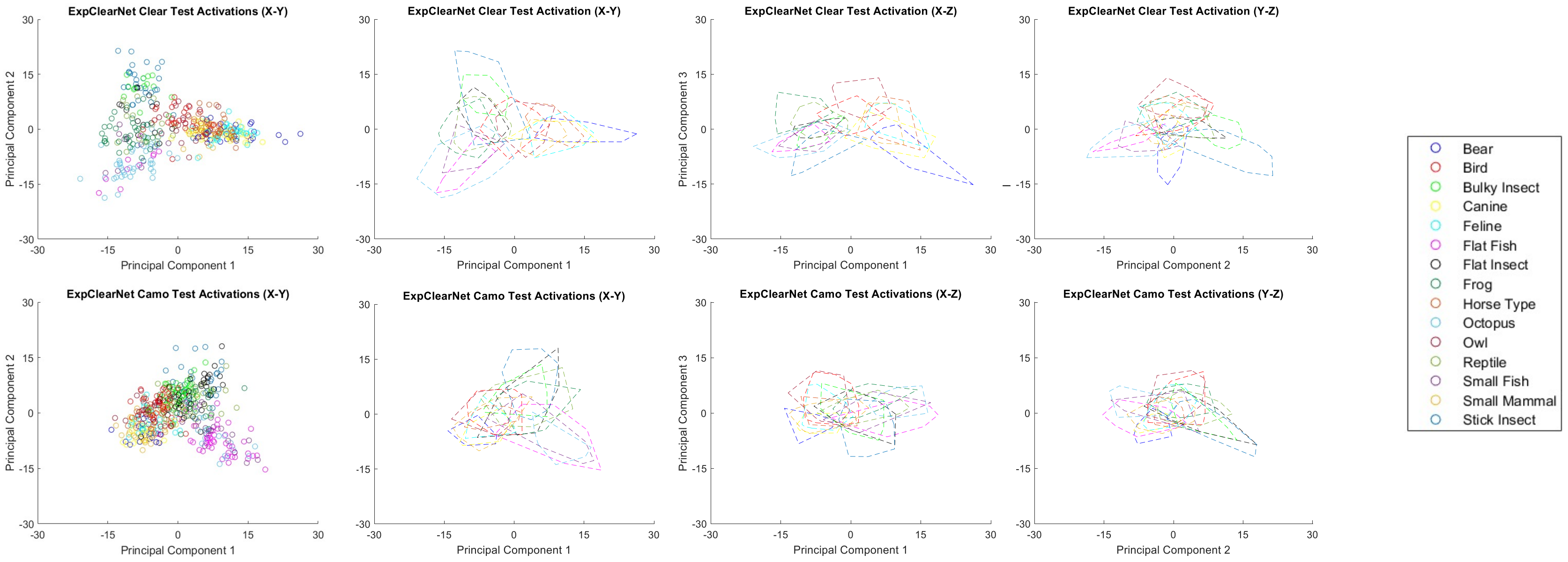

**Figure S4 –** Feature space comparison between ExpClearNet activations on clear and camouflage animal testing images across the first three principal components**.** Conventions are as in Figure 1. **(A)** Cluster representation of clear images on XY axis. **(B-D)** Region representation of clear images on XY, XZ, and YZ axes. **(E)** Cluster representation of camouflage images on XY axis. **(F-H)** Region representation of camouflage images on XY, XZ, and YZ axes.

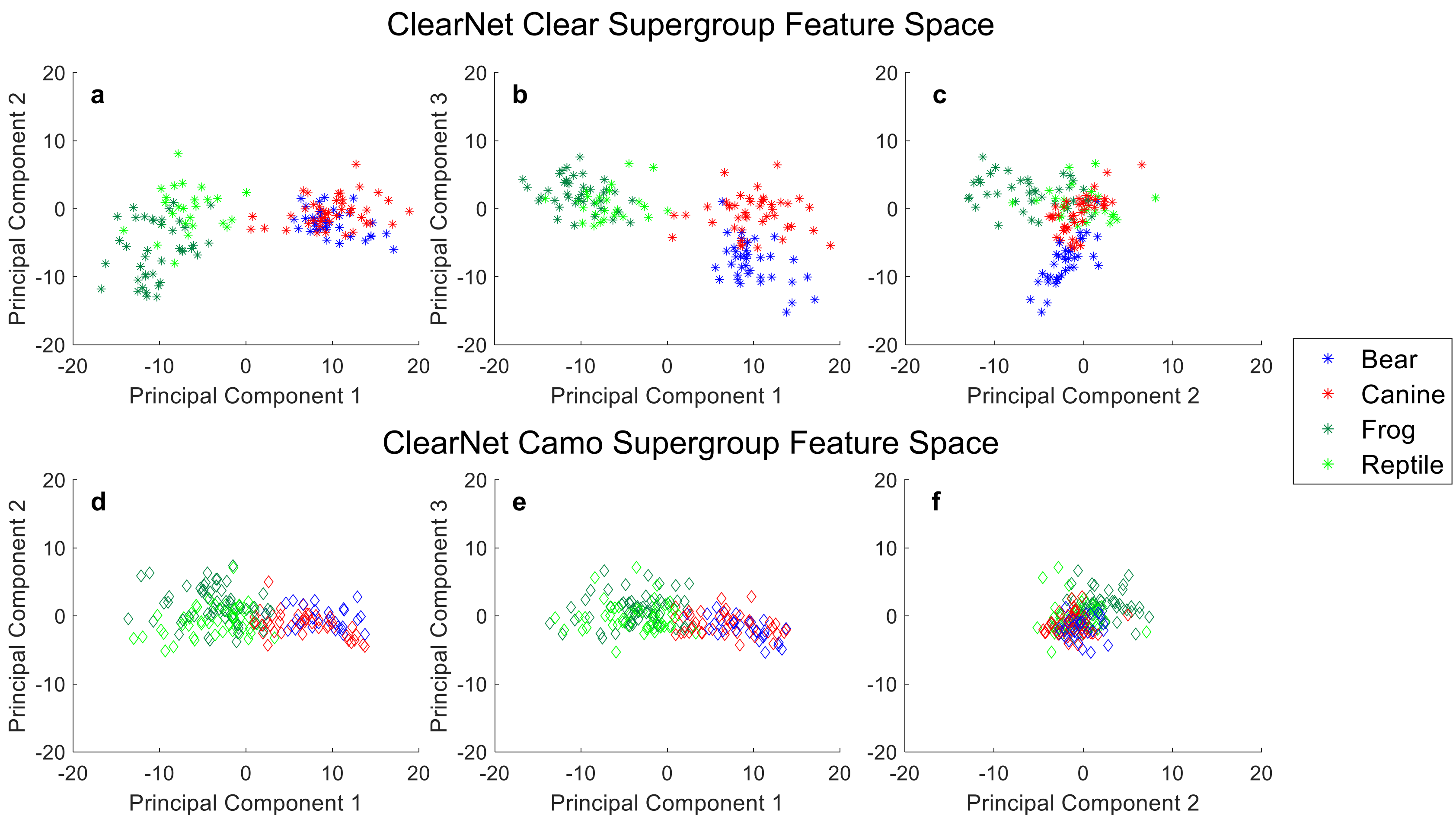

**Figure S5** – Feature space comparison between ClearNet activations on select clear and camouflage animal supergroup images across the first three principal components**.** **(a-c)** Cluster representation of clear images on XY, XZ, and YZ axes. **(d-f)** Cluster representation of camouflage images on XY, XZ, and YZ axes.

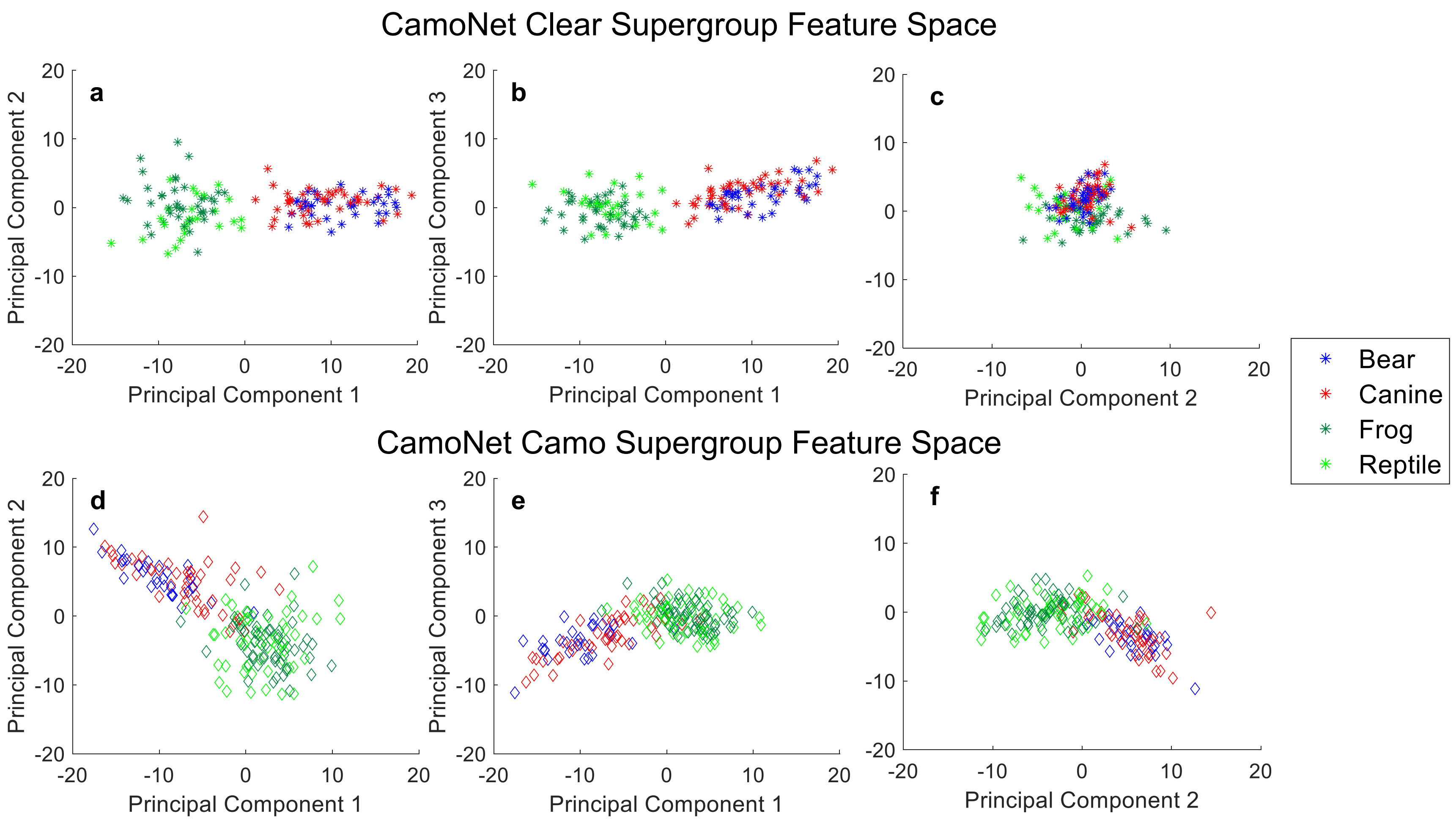

**Figure S6 –** Feature space comparison between CamoNet activations on select clear and camouflage animal supergroup images across the first three principal components. **(a-c)** Cluster representation of clear images on XY, XZ, and YZ axes. **(d-f)** Cluster representation of camouflage images on XY, XZ, and YZ axes.

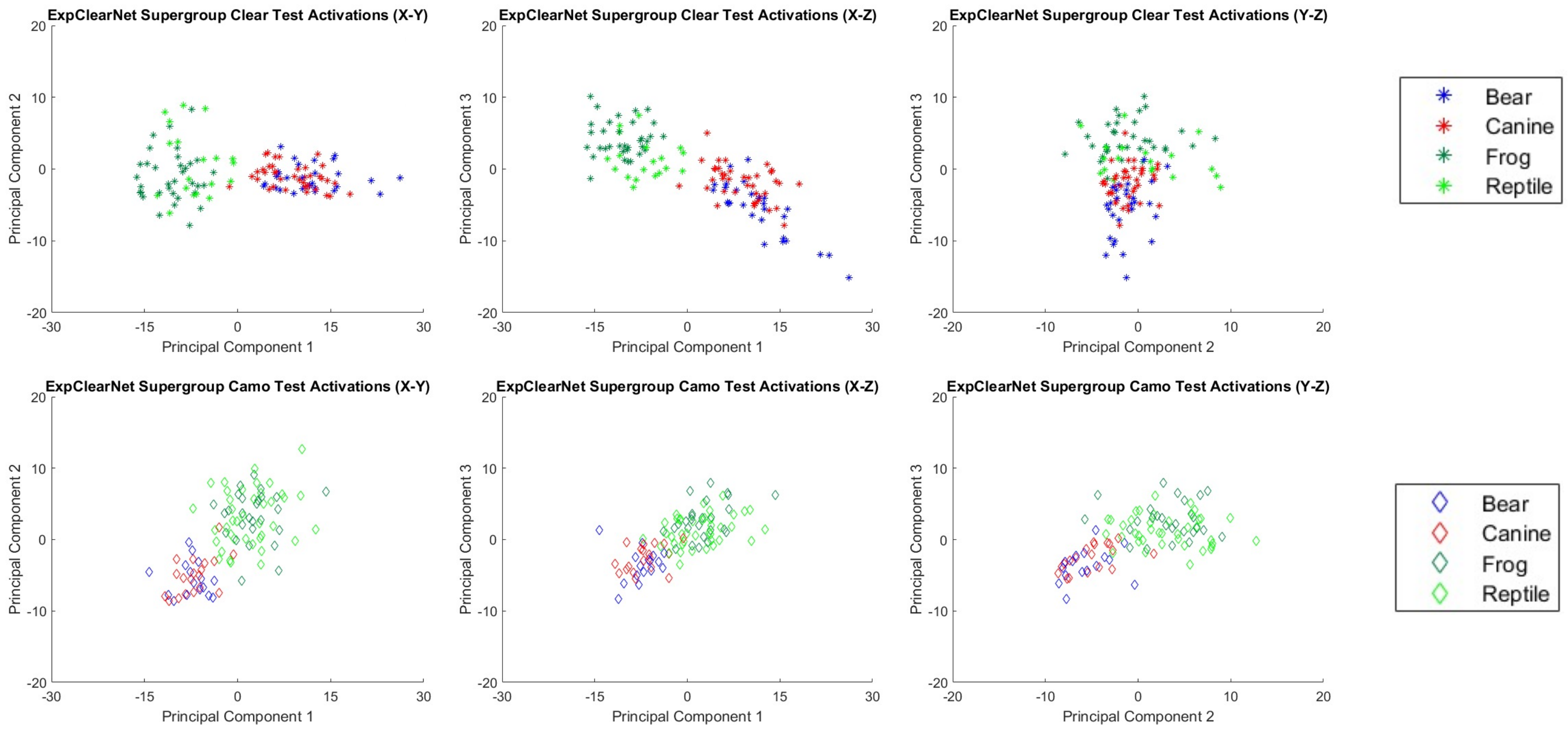

**Figure S7** – Feature space comparison between ExpClearNet activations on select clear and camouflage animal supergroup images across the first three principal components**.** **(a-c)** Cluster representation of clear images on XY, XZ, and YZ axes. **(d-f)** Cluster representation of camouflage images on XY, XZ, and YZ axes.

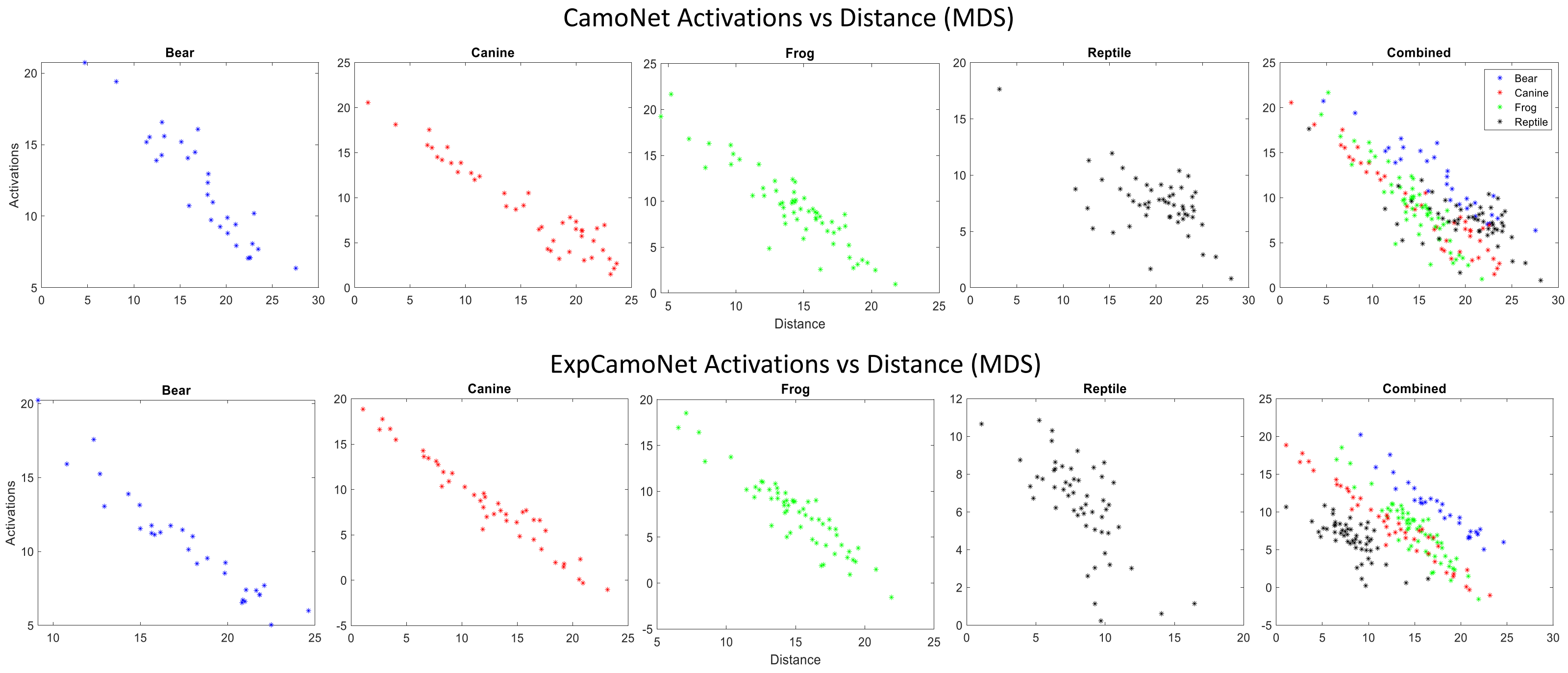

**Figure S8** Comparison of network activation as a function of distance between CamoNet **(a-e)** and ExpCamoNet **(f-j),** generated with Multi-Dimensional Scaling (MDS). Four animal categories for each network are shown individually – Bear **(a,f)**, Canine **(b,g)**, Frog **(c,h)**, and Reptile **(d,i)** – and together **(e,j)**.